\documentclass{article}

\usepackage{arxiv}

\usepackage[utf8]{inputenc} 
\usepackage[T1]{fontenc}    
\usepackage{hyperref}       
\usepackage{url}            
\usepackage{booktabs}       
\usepackage{amsfonts}       
\usepackage{nicefrac}       
\usepackage{microtype}      

\usepackage{lipsum}         
\usepackage{graphicx}
\usepackage{natbib}
\usepackage{doi}
\usepackage{makecell}
\usepackage{amssymb,cuted}
\usepackage{mathtools} 
\usepackage{amsmath} 
\usepackage{cleveref}       
\graphicspath{{./images/}} 
\crefdefaultlabelformat{#2\bfseries\upshape(#1)#3}
\creflabelformat{eq.}{#2\bfseries\upshape(#1)#3}

\def\q{{\boldsymbol q}}
\def\k{{\boldsymbol k}}
\def\v{{\boldsymbol v}}
\def\p{{\boldsymbol p}}
\def\x{{\boldsymbol x}}
\def\W{{\boldsymbol W}}
\def\R{{\boldsymbol R}}
\def\Q{{\boldsymbol Q}}
\def\K{{\boldsymbol K}}
\def\V{{\boldsymbol V}}

\title{RoFormer: Enhanced Transformer with Rotary Position Embedding}

\author{
	Jianlin Su \\
	Zhuiyi Technology Co., Ltd.\\
	Shenzhen \\
	\texttt{bojonesu@wezhuiyi.com} \\
	\And
	Yu Lu \\
	Zhuiyi Technology Co., Ltd.\\
	Shenzhen \\
	\texttt{julianlu@wezhuiyi.com} \\
	\And
	Shengfeng Pan \\
	Zhuiyi Technology Co., Ltd.\\
	Shenzhen \\
	\texttt{nickpan@wezhuiyi.com} \\
	\And
	Ahmed Murtadha \\
	Zhuiyi Technology Co., Ltd.\\
	Shenzhen \\
	\texttt{mengjiayi@wezhuiyi.com} \\
	\And
	Bo Wen \\
	Zhuiyi Technology Co., Ltd.\\
	Shenzhen \\
	\texttt{brucewen@wezhuiyi.com} \\
	\And
	Yunfeng Liu \\
	Zhuiyi Technology Co., Ltd.\\
	Shenzhen \\
	\texttt{glenliu@wezhuiyi.com} \\
}
\renewcommand{\headeright}{}
\renewcommand{\undertitle}{}
\renewcommand{\shorttitle}{RoFormer}

\hypersetup{
pdftitle={RoFormer}
}

\begin{document}
\maketitle

\begin{abstract}
	Position encoding recently has shown effective in the transformer architecture. It enables valuable supervision for dependency modeling between elements at different positions of the sequence. In this paper, we first investigate various methods to integrate positional information into the learning process of transformer-based language models. Then, we propose a novel method named Rotary Position Embedding(RoPE) to effectively leverage the positional information. Specifically, the proposed RoPE encodes the absolute position with a rotation matrix and meanwhile incorporates the explicit relative position dependency in self-attention formulation. Notably, RoPE enables valuable properties, including the flexibility of sequence length, decaying inter-token dependency with increasing relative distances, and the capability of equipping the linear self-attention with relative position encoding.
	Finally, we evaluate the enhanced transformer with rotary position embedding, also called RoFormer, on various long text classification benchmark datasets. Our experiments show that it consistently overcomes its alternatives. Furthermore, we provide a theoretical analysis to explain some experimental results. RoFormer is already integrated into Huggingface: \url{https://huggingface.co/docs/transformers/model_doc/roformer}.
\end{abstract}

\section{Introduction}
The sequential order of words is of great value to natural language understanding. Recurrent neural networks (RRNs) based models encode tokens' order by recursively computing a hidden state along the time dimension.  Convolution neural networks (CNNs) based models (CNNs) \cite{Gehring:2017} were typically considered position-agnostic, but recent work \cite{Islam2020HowMP} has shown that the commonly used padding operation can implicitly learn position information. 
Recently,  the pre-trained language models (PLMs), which were built upon the transformer \cite{Vaswani:2017}, have achieved the state-of-the-art performance of various natural language processing (NLP) tasks, including context representation learning \cite{Devlin2019BERTPO}, machine translation \cite{Vaswani:2017}, and language modeling \cite{Radford2019LanguageMA}, to name a few.  Unlike, RRNs and CNNs-based models, PLMs utilize the self-attention mechanism to semantically capture the contextual representation of a given corpus. As a consequence, PLMs achieve a significant improvement in terms of parallelization over RNNs and improve the modeling ability of longer intra-token relations compared to CNNs\footnote{A stack of multiple CNN layers can also capture longer intra-token relation, here we only consider single layer setting.}. 

It is noteworthy that the self-attention architecture of the current PLMs has shown to be position-agnostic \cite{Yun2020Are}. Following this claim, various approaches have been proposed to encode the position information into the learning process. On one side, generated absolute position encoding through a pre-defined function \cite{Vaswani:2017} was added to the contextual representations, while a trainable absolute position encoding \cite{Gehring:2017,Devlin2019BERTPO,Lan2020ALBERT:2020,clark2020electra,Radford2019LanguageMA,Radford2018ImprovingLU:2018}. On the other side, the previous work \cite{Parikh2016ADA,Shaw2018SelfAttentionWR,Huang2018MusicT,Dai2019TransformerXLAL,Yang2019XLNetGA,Raffel2020ExploringTL,Ke2020RethinkingPE,He2020DeBERTaDB,huang-etal-2020-improve} focuses on relative position encoding, which typically encodes the relative position information into the attention mechanism. In addition to these approaches, the authors of \cite{LiuYDH:2020} have proposed to model the dependency of position encoding from the perspective of Neural ODE \cite{ChenRBD:2018}, and the authors of \cite{Wang2020Encoding} have proposed to model the position information in complex space. Despite the effectiveness of these approaches, they commonly add the position information to the context representation and thus render them unsuitable for the linear self-attention architecture.


In this paper, we introduce a novel method, namely Rotary Position Embedding(RoPE), to leverage the positional information into the learning process of PLMS. Specifically, RoPE encodes the absolute position with a rotation matrix and meanwhile incorporates the explicit relative position dependency in self-attention formulation. Note that the proposed RoPE is prioritized over the existing methods through valuable properties, including the sequence length flexibility, decaying inter-token dependency with increasing relative distances, and the capability of equipping the linear self-attention with relative position encoding.
Experimental results on various long text classification benchmark datasets show that the enhanced transformer with rotary position embedding, namely RoFormer, can give better performance compared to baseline alternatives and thus demonstrates the efficacy of the proposed RoPE.

In brief, our contributions are three-folds as follows:
\begin{itemize}
	\item We investigated the existing approaches to the relative position encoding and found that they are mostly built based on the idea of the decomposition of adding position encoding to the context representations. We introduce a novel method, namely Rotary Position Embedding(RoPE), to leverage the positional information into the learning process of PLMS. The key idea is  to encode relative position by multiplying the context representations with a rotation matrix with a clear theoretical interpretation.
	\item We study the properties of RoPE and show that it decays with the relative distance increased, which is desired for natural language encoding.  We kindly argue that previous relative position encoding-based approaches are not compatible with linear self-attention.
	
	\item We evaluate the proposed RoFormer on various long text benchmark datasets. Our experiments show that it consistently achieves better performance compared to its alternatives.  Some experiments with pre-trained language models are available on GitHub: \url{https://github.com/ZhuiyiTechnology/roformer}. 
\end{itemize}

The remaining of the paper is organized as follows. We establish a formal description of the position encoding problem in self-attention architecture and revisit previous works in \Cref{sec:background}. We then describe the rotary position encoding (RoPE) and study its properties in \Cref{sec:approach}. We report experiments in \Cref{sec:experiment}. Finally, we conclude this paper in \Cref{sec:conclu}.

\section{Background and Related Work}\label{sec:background}
\subsection{Preliminary}
Let $\mathbb{S}_N=\{w_i\}_{i=1}^{N}$ be a sequence of $N$ input tokens with $w_i$ being the $i^{th}$ element. The corresponding word embedding of $\mathbb{S}_N$ is denoted as $\mathbb{E}_N = \{\x_i\}_{i=1}^{N}$, where $\x_i\in \mathbb{R}^{d}$ is the d-dimensional word embedding vector of token $w_i$ without position information. The self-attention first incorporates position information to the word embeddings and transforms them into queries, keys, and value representations. 
\begin{equation}
	\begin{aligned}
		\q_m &=f_q(\x_m, m)\\
		\k_n &=f_k(\x_n, n)\\
		\v_n &=f_v(\x_n, n),\\
	\end{aligned}
	\label{fn:qkv}
\end{equation}
where $\q_m,\k_n$ and $ \v_n$ incorporate the $m^{th}$ and $n^{th}$ positions through $f_q,f_k$ and $f_v$, respectively. The query and key values are then used to compute the attention weights, while  the output is computed as the weighted sum over the value representation. 
\begin{equation}
	\begin{aligned}
		a_{m,n}&=\frac{\exp(\frac{\q_m^{\intercal}\k_n}{\sqrt{d}})}{\sum_{j=1}^{N}\exp(\frac{\q_m^{\intercal}\k_j}{\sqrt{d}})}\\
		\mathbf{o}_m&=\sum_{n=1}^{N}a_{m,n}\v_{n}
	\end{aligned}
	\label{fn:attn}
\end{equation}
The existing approaches of transformer-based position encoding mainly focus on choosing a suitable function to form \Cref{fn:qkv}.

\subsection{Absolute position embedding}
A typical choice of \Cref{fn:qkv} is 
\begin{equation}
	f_{t:t\in\{q,k,v\}}(\x_i,i):=\W_{t:t\in\{q,k,v\}}(\x_i+\p_i),
	\label{fn:adtv-posi}
\end{equation}
where $\p_i\in\mathbb{R}^{d}$ is a d-dimensional vector depending of the position of token $\x_i$. Previous work \cite{Devlin2019BERTPO,Lan2020ALBERT:2020,clark2020electra,Radford2019LanguageMA,Radford2018ImprovingLU:2018} introduced the use of a set of trainable vectors $\p_i\in\{\p_t\}_{t=1}^{L}$, where $L$ is the maximum sequence length. The authors of \cite{Vaswani:2017} have proposed to generate $\p_i$ using the sinusoidal function.
\begin{equation}
	\begin{cases}
		\p_{i,2t}&=\sin(k/10000^{2t/d})\\
		\p_{i,2t+1}&=\cos(k/10000^{2t/d})
	\end{cases}
	\label{fn:sins}
\end{equation}
in which $\p_{i,2t}$ is the $2t^{th}$ element of the d-dimensional vector $\p_i$. In the next section, we show that our proposed RoPE is related to this intuition from the sinusoidal function perspective. However, instead of directly adding the position to the context representation, RoPE proposes to incorporate the relative position information by multiplying with the sinusoidal functions.

\subsection{Relative position embedding}
The authors of \cite{Shaw2018SelfAttentionWR} applied  different settings of \Cref{fn:qkv} as following:
\begin{equation}
	\begin{aligned}
		f_q(\x_m):=\W_{q}\x_m\\
		f_k(\x_n, n):=\W_{k}(\x_n+\tilde{\p}^k_r)\\
		f_v(\x_n, n):=\W_{v}(\x_n+\tilde{\p}^v_r)\\
	\end{aligned}
	\label{fn:shaw}
\end{equation}
where $\tilde{\p}^k_r,\tilde{\p}^v_r\in\mathbb{R}^{d}$ are trainable relative position embeddings. Note that $r=\operatorname{clip}(m-n,r_{\text{min}},r_{\text{max}})$ represents the relative distance between position $m$ and $n$. They clipped the relative distance with the hypothesis that precise relative position information is not useful beyond a certain distance.
Keeping the form of \Cref{fn:adtv-posi}, the authors \cite{Dai2019TransformerXLAL} have proposed to decompose $\q_m^{\intercal}\k_n$ of \Cref{fn:attn} as 
\begin{equation}
	\q_m^{\intercal}\k_n=\x_m^{\intercal}\W_q^{\intercal}\W_k\x_n+\x_m^{\intercal}\W_q^{\intercal}\W_k\p_n+\p_m^{\intercal}\W_q^{\intercal}\W_k\x_n+\p_m^{\intercal}\W_q^{\intercal}\W_k\p_n,
	\label{fn:rela-posi1}
\end{equation}
the key idea is to replace the absolute position embedding $\p_n$ with its sinusoid-encoded relative counterpart $\tilde{\p}_{m-n}$, while the absolute position $\p_m$ in the third and fourth term with two trainable vectors $\mathbf{u}$ and $\mathbf{v}$ independent of the query positions. Further, $\W_k$ is distinguished for the content-based and location-based key vectors $\x_n$ and $\p_n$, denoted as $\W_k$ and $\widetilde{\W}_k$, resulting in: 
\begin{equation}
	\q_m^{\intercal}\k_n=\x_m^{\intercal}\W_q^{\intercal}\W_k\x_n+\x_m^{\intercal}\W_q^{\intercal}\widetilde{\W}_k\tilde{\p}_{m-n}+\mathbf{u}^{\intercal}\W_q^{\intercal}\W_k\x_n+\mathbf{v}^{\intercal}\W_q^{\intercal}\widetilde{\W}_k\tilde{\p}_{m-n}
	\label{fn:rela-posi2}
\end{equation}

It is noteworthy that the position information in the value term is removed by setting $f_v(\x_j):=\W_{v}\x_j$. Later work \cite{Raffel2020ExploringTL,He2020DeBERTaDB,Ke2020RethinkingPE,huang-etal-2020-improve} followed these settings by only encoding the relative position information into the attention weights. However, the authors of \cite{Raffel2020ExploringTL} reformed \Cref{fn:rela-posi1} as: 
\begin{equation}
	\q_m^{\intercal}\k_n=\x_m^{\intercal}\W_q^{\intercal}\W_k\x_n + b_{i,j}
	\label{fn:rela-posi3}
\end{equation}
where $b_{i,j}$ is a trainable bias. The authors of \cite{Ke2020RethinkingPE} investigated the middle two terms of \Cref{fn:rela-posi1} and found little correlations between absolute positions and words. The authors of \cite{Raffel2020ExploringTL} proposed to model a pair of words or positions using different projection matrices.
\begin{equation}
	\q_m^{\intercal}\k_n=\x_m^{\intercal}\W_q^{\intercal}\W_k\x_n +\p_m^{\intercal}\mathbf{U}_q^{\intercal}\mathbf{U}_k\p_n+b_{i,j}
	\label{fn:rela-posi4}
\end{equation}
The authors of \cite{He2020DeBERTaDB} argued that the relative positions of two tokens could only be fully modeled using the middle two terms of \Cref{fn:rela-posi1}. As a consequence, the absolute position embeddings $\p_m$ and $\p_n$ were simply replaced with the relative position embeddings $\tilde{\p}_{m-n}$:
\begin{equation}
	\q_m^{\intercal}\k_n=\x_m^{\intercal}\W_q^{\intercal}\W_k\x_n+\x_m^{\intercal}\W_q^{\intercal}\W_k\tilde{\p}_{m-n}+\tilde{\p}_{m-n}^{\intercal}\W_q^{\intercal}\W_k\x_n
	\label{fn:debeta}
\end{equation}
A comparison of the four variants of the relative position embeddings \cite{Radford2018ImprovingLU:2018} has shown that the variant similar to \Cref{fn:debeta} is the most efficient among the other three. Generally speaking, all these approaches attempt to  modify \Cref{fn:rela-posi1} based on the decomposition of \Cref{fn:adtv-posi} under the self-attention settings in \Cref{fn:attn}, which was originally proposed in \cite{Vaswani:2017}. They commonly introduced to directly add the position information to the context representations. Unlikely, our approach aims to derive the relative position encoding from \Cref{fn:qkv} under some constraints. Next, we show that the derived approach is more interpretable by incorporating relative position information with the rotation of context representations. 
\section{Proposed approach}
\label{sec:approach}
In this section, we discuss the proposed rotary position embedding (RoPE). We first formulate the relative position encoding problem in  \Cref{sec:formulation}, we then derive the RoPE in \Cref{sec:RoPE} and investigate its properties in  \Cref{sec: prop of RoPE}. 

\subsection{Formulation}
\label{sec:formulation}
Transformer-based language modeling usually leverages the position information of individual tokens through a self-attention mechanism. As can be observed in \Cref{fn:attn}, $\q_m^{\intercal}\k_n$ typically enables knowledge conveyance between tokens at different positions. In order to incorporate relative position information, we require the inner product of query $\q_m$ and key $\k_n$ to be formulated by a function $g$, which takes only the word embeddings $\x_m$, $\x_n$, and their relative position $m -n$ as input variables. In other words, we hope that the inner product encodes position information only in the relative form:

\if We start with \cref{fn:qkv}. To incorporate position information with self-attention in \cref{fn:attn}, we first consider incorporating the position information in the query and key, respectively.  Furthermore, we set the inner product of these two terms to a function explicitly depending on their relative distance. In other words, we hope the inner product only encodes relative position. \fi


\begin{equation}
	\langle f_q(\x_m, m),f_k(\x_n, n)\rangle=g(\x_m,\x_n,m-n).
	\label{fn:formulation}
\end{equation}
The ultimate goal is to find an equivalent encoding mechanism to solve the functions $f_q(\x_m, m)$ and $f_k(\x_n, n)$ to conform the aforementioned relation. 

\subsection{Rotary position embedding}
\label{sec:RoPE}

\subsubsection{A 2D case}
We begin with a simple case with a dimension $d=2$. Under these settings, we make use of the geometric property of vectors on a 2D plane and its complex form to prove (refer \Cref{appendix:rope-deriv} for more details) that a solution to our formulation \Cref{fn:formulation} is: 

\begin{equation}
	\begin{aligned}
		f_q(\x_m, m) &= (\W_q\x_m)e^{im\theta}\\
		f_k(\x_n, n) &= (\W_k\x_n)e^{in\theta}\\
		g(\x_m, \x_n, m - n) &= \operatorname{Re}[(\W_q\x_m)(\W_k\x_n)^{*}e^{i(m - n)\theta}]
	\end{aligned}
	\label{fn:res-2d}
\end{equation}
where $\operatorname{Re}[\cdot]$ is the real part of a complex number and $(\W_k\x_n)^{*}$ represents the conjugate complex number of $(\W_k\x_n)$. $\theta \in \mathbb{R}$ is a preset non-zero constant. We can further write $f_{\{q, k\}}$ in a multiplication matrix: 
\begin{equation}
	f_{\{q, k\}}(\x_m, m) = \left(
	\begin{array}{cc}
		\cos{m\theta}& -\sin{m\theta}  \\
		\sin{m\theta}&\cos{m\theta} 
	\end{array}
	\right)
	\left(
	\begin{array}{cc}
		W^{(11)}_{\{q, k\}} & W^{(12)}_{\{q, k\}} \\
		W^{(21)}_{\{q, k\}} & W^{(22)}_{\{q, k\}}
	\end{array}
	\right)
	\left(
	\begin{array}{cc}
		x^{(1)}_m\\
		x^{(2)}_m
	\end{array}
	\right)
\end{equation}
where $(x^{(1)}_m, x^{(2)}_m)$ is $\x_m$ expressed in the 2D coordinates. Similarly, $g$ can be viewed as a matrix  and thus enables the solution of formulation in \Cref{sec:formulation} under the 2D case. Specifically, incorporating the relative position embedding is straightforward: simply rotate the affine-transformed word embedding vector by amount of angle multiples of its position index and thus interprets the intuition behind \textit{Rotary Position Embedding}. 

\subsubsection{General form}\label{subsec:general-form}
\begin{figure}[hb]
	\centering
	\includegraphics[width=0.7\textwidth]{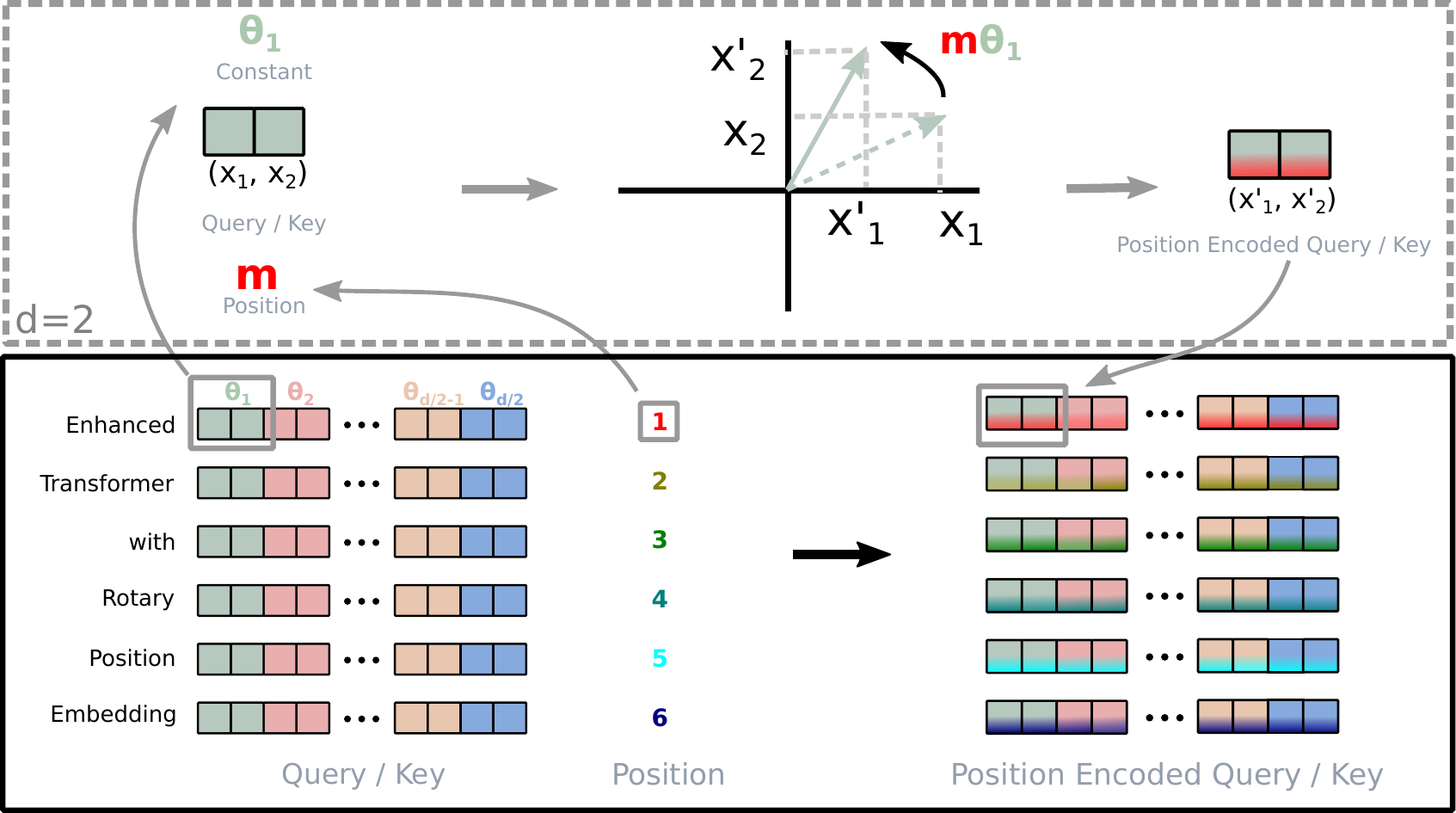}
	\caption{Implementation of Rotary Position Embedding(RoPE). }
	\label{fig:rope}
\end{figure}

In order to generalize our results in 2D to any $\x_i \in \mathbb{R}^d$ where $d$ is even,  we divide the d-dimension space into $d/2$ sub-spaces and combine them in the merit of the linearity of the inner product, turning $f_{\{q, k\}}$ into: 

\begin{equation}
	f_{\{q, k\}}(\x_m, m) = \R^d_{\Theta, m}\W_{\{q, k\}}\x_m 
	\label{fn:rope-fqk}
\end{equation}
where 
\begin{equation}    
	\R^d_{\Theta,m} = 
	\begin{pmatrix}
		\cos{m\theta_1}& -\sin{m\theta_1}&0&0&\cdots&0&0\\
		\sin{m\theta_1}&\cos{m\theta_1}&0&0&\cdots&0&0 \\
		0&0&\cos{m\theta_2}& -\sin{m\theta_2}&\cdots&0&0\\
		0&0&\sin{m\theta_2}&\cos{m\theta_2}&\cdots&0&0 \\
		\vdots&\vdots&\vdots&\vdots&\ddots&\vdots&\vdots\\
		0&0&0&0&\cdots&\cos{m\theta_{d/2}}& -\sin{m\theta_{d/2}}\\
		0&0&0&0&\cdots&\sin{m\theta_{d/2}}&\cos{m\theta_{d/2}}
	\end{pmatrix}
	\label{fn:rope-RMat}
\end{equation}
is the rotary matrix with pre-defined parameters $\Theta = \{\theta_i=10000^{-2(i-1)/d}, i \in [1, 2, ..., d/2]\}$. A graphic illustration of RoPE is shown in \Cref{fig:rope}. Applying our RoPE to self-attention in \Cref{fn:attn}, we obtain:
\begin{equation}
	\q_m^{\intercal}\k_n 
	=(\R^d_{\Theta, m}\W_q\x_m)^\intercal(\R^d_{\Theta, n}\W_k\x_n) =\x^\intercal\W_qR^d_{\Theta, n-m}\W_k\x_n
	\label{fn:rope-qk}
\end{equation}
where $\R^d_{\Theta, n-m}=(\R^{d}_{\Theta, m})^\intercal\R^d_{\Theta, n}$. Note that $\R^d_\Theta$ is an orthogonal matrix, which ensures stability during the process of encoding position information. In addition, due to the sparsity of $R^d_\Theta$, applying matrix multiplication directly as in \Cref{fn:rope-qk} is not computationally efficient; we provide another realization in theoretical explanation.

In contrast to the additive nature of position embedding method adopted in the previous works, i.e.,  \Cref{fn:adtv-posi,fn:sins,fn:shaw,fn:rela-posi1,fn:rela-posi2,fn:rela-posi3,fn:rela-posi4,fn:debeta}, our approach is multiplicative. Moreover,  RoPE naturally incorporates relative position information through rotation matrix product instead of altering terms in the expanded formulation of additive position encoding when applied with self-attention. 

\subsection{Properties of RoPE}
\label{sec: prop of RoPE}
\paragraph{Long-term decay:} 
Following \cite{Vaswani:2017}, we set $\theta_i=10000^{-2i/d}$. One can prove that this setting provides a long-term decay property (refer to \Cref{appendix:long-term-decay} for more details), which means the inner-product will decay when the relative position increase. This property coincides with the intuition that a pair of tokens with a long relative distance should have less connection.

\paragraph{RoPE with linear attention:}
The self-attention can be rewritten in a more general form.

\begin{equation}
	\operatorname{Attention}(\mathbf{Q},\mathbf{K},\mathbf{V})_m=\frac{\sum_{n=1}^{N}\operatorname{sim}(\q_m,\k_n)\v_n}{\sum_{n=1}^{N}\operatorname{sim}(\q_m, \k_n)}.
	\label{fn:atten-full}
\end{equation}
The original self-attention chooses $\operatorname{sim}(\q_m,\k_n)=\exp(\q_m^{\intercal}\k_n/\sqrt{d})$. Note that the original self-attention should compute the inner product of query and key for every pair of tokens, which has a quadratic complexity $\mathbb{O}(N^2)$.  Follow \cite{katharopoulos2020transformers}, the linear attentions reformulate \Cref{fn:atten-full} as 

\begin{equation}
	\operatorname{Attention}(\Q,\K,\V)_m=\frac{\sum_{n=1}^{N}\phi(\q_m)^{\intercal}\varphi(\k_n)\v_n}{\sum_{n=1}^{N}\phi(\q_m)^{\intercal}\varphi(\k_n)},
	\label{fn:linear-attention}
\end{equation}
where $\phi(\cdot), \varphi(\cdot)$ are usually non-negative functions. The authors of \cite{katharopoulos2020transformers} have proposed $\phi(x)=\varphi(x)=\operatorname{elu}(x)+1$ and first computed the multiplication between keys and values using the associative property of matrix multiplication.  A softmax function is used in \cite{shen2021efficient} to normalize queries and keys separately before the inner product, which is equivalent to   $\phi(\q_i)=\operatorname{softmax}(\q_i)$ and $\phi(\k_j)=\exp(\k_j)$. For more details about linear attention, we encourage readers to refer to original papers. In this section, we focus on discussing incorporating RoPE with \Cref{fn:linear-attention}. Since RoPE injects position information by rotation, which keeps the norm of hidden representations unchanged, we can combine RoPE with linear attention by multiplying the rotation matrix with the outputs of the non-negative functions.

\begin{equation}
	\operatorname{Attention}(\mathbf{Q},\mathbf{K},\mathbf{V})_m=\frac{\sum_{n=1}^{N}\big(\R^d_{\Theta, m}\phi(\q_m)\big)^{\intercal}\big(\R^d_{\Theta, n}\varphi(\k_n)\big)\v_n}{\sum_{n=1}^{N}\phi(\q_m)^{\intercal}\varphi(\k_n)}.
	\label{fn:linear-rope}
\end{equation}

It is noteworthy that we keep the denominator unchanged to avoid the risk of dividing zero, and the summation in the numerator could contain negative terms. Although the weights for each value $\v_i$ in  \Cref{fn:linear-rope} are not strictly probabilistic normalized, we kindly argue that the computation can still model the importance of values. 
\subsection{Theoretical Explanation}
\subsubsection{Derivation of RoPE under 2D}\label{appendix:rope-deriv}
Under the case of $d=2$, we consider two-word embedding vectors $\x_q$, $\x_k$ corresponds to query and key and their position $m$ and $n$, respectively. According to \cref{fn:qkv}, their position-encoded counterparts are: 
\begin{equation}
	\begin{aligned}
		\q_m &= f_q(\x_q, m),  \\
		\k_n &= f_k(\x_k, n),
	\end{aligned}
	\label{fn:deri-qk}
\end{equation}
where the subscripts of $\q_m$ and $\k_n$ indicate the encoded positions information. Assume that there exists a function $g$ that defines the inner product between vectors produced by $f_{\{q, k\}}$: 
\begin{equation}
	\q^\intercal_m\k_n = \langle f_q(\x_m, m),f_k(\x_n, n)\rangle= g(\x_m, \x_n, n - m),
	\label{fn:deri-g}
\end{equation}
we further require below initial condition to be satisfied:
\begin{equation}
	\begin{aligned}
		\q &= f_q(\x_q, 0), \\
		\k &= f_k(\x_k, 0), 
	\end{aligned}
	\label{fn:deri-initCond1}
\end{equation}
which can be read as the vectors with empty position information encoded. Given these settings, we attempt to find a solution of $f_q$, $f_k$.  First, we take advantage of the geometric meaning of vector in 2D and its complex counter part, decompose functions in \Cref{fn:deri-qk,fn:deri-g} into:
\begin{equation}
	\begin{aligned}
		f_q(\x_q, m) &= R_q(\x_q, m)e^{i\Theta_q(\x_q, m)}, \\
		f_k(\x_k, n) &= R_k(\x_k, n)e^{i\Theta_k(\x_k, n)}, \\
		g(\x_q, \x_k, n - m) &= R_g(\x_q, \x_k, n - m)e^{i\Theta_g(\x_q, \x_k, n - m)},
	\end{aligned}
\end{equation}
where $R_f$, $R_g$ and $\Theta_f$, $\Theta_g$ are the radical and angular components for $f_{\{q, k\}}$ and $g$, respectively. Plug them into \Cref{fn:deri-g}, we get the relation: 

\begin{equation}
	\begin{aligned}
		R_q(\x_q, m)R_k(\x_k, n) &= R_g(\x_q, \x_k, n - m), \\
		\Theta_k(\x_k, n) - \Theta_q(\x_q, m) &= \Theta_g(\x_q, \x_k, n - m),
	\end{aligned}
	\label{fn:deri-rtheta}
\end{equation}
with the corresponding initial condition as: 
\begin{equation}
	\begin{aligned}
		\q &= \|\q\|e^{i\theta_q} = R_q(\x_q, 0)e^{i\Theta_q(\x_q, 0)},\\
		\k & = \|\k\|e^{i\theta_k} = R_k(\x_k, 0)e^{i\Theta_k(\x_k, 0)},
	\end{aligned}
	\label{fn:deri-initCond2}
\end{equation}
where $\|\q\|$, $\|\k\|$ and $\theta_q$, $\theta_k$ are the radial and angular part of $\q$ and $\k$ on the 2D plane.

Next, we set $m=n$ in \Cref{fn:deri-rtheta} and take into account initial conditions in \Cref{fn:deri-initCond2}: 
\begin{subequations}
	\begin{align}
		R_q(\x_q, m)R_k(\x_k, m) &= R_g(\x_q, \x_k, 0) = R_q(\x_q, 0)R_k(\x_k, 0)  = \|\q\|\|\k\| \label{fn:deri-solveR},\\
		\Theta_k(\x_k, m) - \Theta_q(\x_q, m) &= \Theta_g(\x_q, \x_k, 0) =\Theta_k(\x_k, 0) - \Theta_q(\x_q, 0) =  \theta_k - \theta_q \label{fn:deri-solveTheta}.
	\end{align}
\end{subequations}
On one hand, from, a straightforward solution of $R_f$ could be formed from \Cref{fn:deri-solveR} : 
\begin{equation}
	\begin{aligned}
		R_q(\x_q, m) &= R_q(\x_q, 0) =  \|\q\| \\
		R_k(\x_k, n) &= R_k(\x_k, 0) = \|\k\|\\
		R_g(\x_q, \x_k, n - m) &= R_g(\x_q, \x_k, 0) =  \|\q\|\|\k\|
	\end{aligned}
	\label{fn:deri-solR1}
\end{equation}
which interprets the radial functions $R_q$, $R_k$ and $R_g$ are independent from the position information. On the other hand, as can be noticed in \Cref{fn:deri-solveTheta},  $\Theta_q(\x_q, m) - \theta_q = \Theta_k(\x_k, m) - \theta_k$ indicates that the angular functions does not dependent on query and key, we set them to  $\Theta_f := \Theta_q = \Theta_k$ and term $\Theta_f(\x_{\{q, k\}}, m) - \theta_{\{q, k\}}$ is a function of position $m$ and is independent of word embedding $\x_{\{q, k\}}$, we denote it as $\phi(m)$, yielding:
\begin{equation}
	\Theta_f(\x_{\{q, k\}}, m) = \phi(m) + \theta_{\{q,k\}},
	\label{fn:deri-solTheta1}
\end{equation}
Further, by plugging  $n = m + 1$ to \Cref{fn:deri-rtheta} and consider the above equation, we can get: 
\begin{equation}
	\phi(m + 1) - \phi(m) = \Theta_g(\x_q, \x_k, 1) + \theta_q - \theta_k,
	\label{fn:deri-solTheta2}
\end{equation}
Since RHS is a constant irrelevant to $m$, $\phi(m)$ with continuous integer inputs produce an arithmetic progression:
\begin{equation}
	\phi(m) = m\theta + \gamma,
	\label{fn:deri-solTheta3}
\end{equation}
where $\theta,\gamma \in \mathbb{R}$ are constants and $\theta$ is non-zero. To summarize our solutions from \Cref{fn:deri-solR1,fn:deri-solTheta1,fn:deri-solTheta2,fn:deri-solTheta3}:

\begin{equation}
	\begin{aligned}
		f_q(\x_q, m) &= \|\q\|e^{i\theta_q + m\theta + \gamma} = \q e^{i(m\theta + \gamma)}, \\
		f_k(\x_k, n) &= \|\k\|e^{i\theta_k + n\theta + \gamma}= \k e^{i(n\theta + \gamma)}.
	\end{aligned} 
	\label{fn:deri-solution1}
\end{equation}

Note that we do not apply any constrains to $f_q$ and $f_k$ of \Cref{fn:deri-initCond1}, thus $f_q(\x_m, 0)$ and $f_k(\x_n, 0)$ are left to choose freely. To make our results comparable to \Cref{fn:adtv-posi}, we define:
\begin{equation}
	\begin{aligned}
		\q = f_q(\x_m, 0) &= \W_q\x_n,\\
		\k = f_k(\x_n, 0) &= \W_k\x_n.
	\end{aligned}
\end{equation}
Then, we simply set $\gamma = 0$ in \Cref{fn:deri-solution1} of the final solution: 
\begin{equation}
	\begin{aligned}
		f_q(\x_m, m) &= (\W_q\x_m)e^{im\theta},\\
		f_k(\x_n, n) &= (\W_k\x_n)e^{in\theta}.
	\end{aligned}
\end{equation}

\subsubsection{Computational efficient realization of rotary matrix multiplication}\label{appendix:rope-efficient}
Taking the advantage of the sparsity of $\R^d_{\Theta, m}$ in \Cref{fn:rope-RMat}, a more computational efficient realization of a multiplication of $R^d_\Theta$ and $\x \in \mathbb{R}^d$ is:  
\begin{equation}
	\R^d_{\Theta, m}\x = 
	\begin{pmatrix}
		x_1\\
		x_2\\
		x_3\\
		x_4\\
		\vdots\\
		x_{d-1}\\
		x_d
	\end{pmatrix}
	\otimes
	\begin{pmatrix}
		\cos{m\theta_1} \\
		\cos{m\theta_1} \\
		\cos{m\theta_2} \\
		\cos{m\theta_2} \\
		\vdots \\
		\cos{m\theta_{d/2}} \\
		\cos{m\theta_{d/2}} 
	\end{pmatrix}
	+
	\begin{pmatrix}
		-x_2\\
		x_1\\
		-x_4\\
		x_3\\
		\vdots\\
		-x_d\\
		x_{d-1}
	\end{pmatrix}
	\otimes
	\begin{pmatrix}
		\sin{m\theta_1}\\
		\sin{m\theta_1}\\
		\sin{m\theta_2}\\
		\sin{m\theta_2}\\
		\vdots\\
		\sin{m\theta_{d/2}}\\
		\sin{m\theta_{d/2}}
	\end{pmatrix}
\end{equation}

\subsubsection{Long-term decay of RoPE}\label{appendix:long-term-decay}
We can group entries of vectors $\q=\W_q\x_m$ and $\k=\W_k\x_n$ in pairs, and the inner product of RoPE in \Cref{fn:rope-qk} can be written as a complex number multiplication.

\begin{equation}
	(\R^d_{\Theta, m}\W_q\x_m)^\intercal(\R^d_{\Theta, n}\W_k\x_n) = \operatorname{Re}\bigg[\sum_{i=0}^{d/2-1}\q_{[2i:2i+1]}\k_{[2i:2i+1]}^{*}e^{i(m-n)\theta_{i}}\bigg]
\end{equation}
where $\q_{[2i:2i+1]}$ represents the $2i^{th}$ to $(2i+1)^{th}$ entries of $\q$. Denote $h_i=\q_{[2i:2i+1]}\k_{[2i:2i+1]}^{*}$ and $S_j=\sum_{i=0}^{j-1}e^{i(m-n)\theta_i}$, and let $h_{d/2}=0$ and $S_0=0$, we can rewrite the summation using Abel transformation

\begin{equation}
	\sum_{i=0}^{d/2-1}\q_{[2i:2i+1]}\k_{[2i:2i+1]}^{*}e^{i(m-n)\theta_{i}}=\sum_{i=0}^{d/2-1}h_i(S_{i+1}-S_{i})=-\sum_{i=0}^{d/2-1}S_{i+1}(h_{i+1}-h_i).
\end{equation}

\begin{figure}
	\centering
	\includegraphics[width=0.7\textwidth]{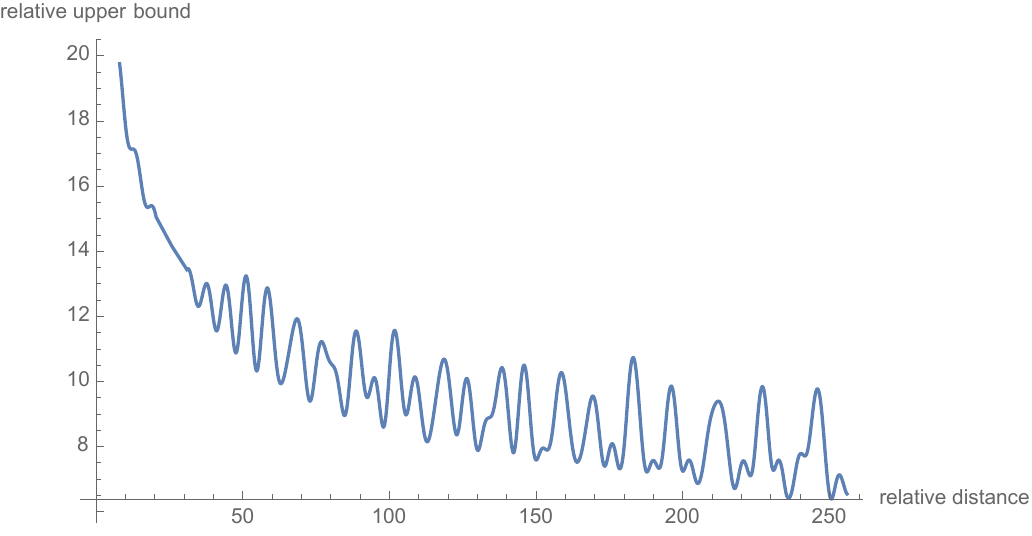}
	\caption{Long-term decay of RoPE.}
	\label{fig:decay}
\end{figure}

Thus,

\begin{equation}
	\begin{aligned}
		\bigg\vert\sum_{i=0}^{d/2-1}\q_{[2i:2i+1]}\k_{[2i:2i+1]}^{*}e^{i(m-n)\theta_{i}}\bigg\vert &= \bigg\vert\sum_{i=0}^{d/2-1}S_{i+1}(h_{i+1}-h_i)\bigg\vert\\
		&\leq \sum_{i=0}^{d/2-1}\vert S_{i+1}\vert\vert(h_{i+1}-{h_i})\vert\\
		&\leq \big(\max_i\vert h_{i+1}-h_i\vert\big)\sum_{i=0}^{d/2-1}\vert S_{i+1}\vert
	\end{aligned}
\end{equation}
Note that the value of $\frac{1}{d/2}\sum_{i=1}^{d/2}\vert S_i\vert$ decay with the relative distance $m-n$ increases by setting $\theta_i=10000^{-2i/d}$, as shown in \Cref{fig:decay}.

\section{Experiments and Evaluation}
\label{sec:experiment}

We evaluate the proposed RoFormer on various NLP tasks as follows. We validate the performance of the proposed solution on machine translation task \Cref{subsec:mt}. Then, we compare our RoPE implementation with BERT\cite{Devlin2019BERTPO} during the pre-training stage in \Cref{subsec:pretrain}. Based on the pre-trained model, in \Cref{subsec:glue}, we further carry out evaluations across different downstream tasks from GLUE benchmarks\cite{glue}. In Addition, we conduct experiments using the proposed RoPE with the linear attention of PerFormer \cite{performer} in \Cref{subsec:peformer}. By the end, additional tests on Chinese data are included in \Cref{subsec:chinese}. All the experiments were run on two cloud severs with 4 x V100 GPUs.

\subsection{Machine Translation}\label{subsec:mt}
We first demonstrate the performance of RoFormer on sequence-to-sequence language translation tasks. 
\subsubsection{Experimental Settings} 
We choose the standard WMT 2014 English-German dataset\cite{mtdataset}, which consists of approximately 4.5 million sentence pairs. We compare to the transformer-based baseline alternative \cite{Vaswani:2017}. 
\subsubsection{Implementation details} 
We carry out some modifications on  self-attention layer of the baseline model \cite{Vaswani:2017} to enable RoPE to its learning process. We replicate the setup for English-to-German translation with a vocabulary of 37k based on a joint source and target byte pair encoding(BPE)\cite{bpe}. During the evaluation, a single model is obtained by averaging the last 5 checkpoints. The result uses beam search with a beam size of 4 and length penalty 0.6.  We implement the experiment in PyTorch in the fairseq toolkit (MIT License)\cite{fairseq}.  Our model is optimized with the Adam optimizer using $\beta_1 = 0.9$, $\beta_2 = 0.98$, learning rate is increased linearly from $1e-7$ to $5e-4$ and then decayed proportionally to the inverse square root of the step number. Label smoothing with 0.1 is also adopted. We report the BLEU\cite{bleu} score on the test set as the final metric.

\subsubsection{Results} 
We train the baseline model and our RoFormer under the same settings and report the results in \Cref{tb:mt}. As can be seen, our model gives better BLEU scores compared to the baseline Transformer. 

\begin{table}
	\caption{The proposed RoFormer gives better BLEU scores compared to its baseline alternative \cite{Vaswani:2017} on the WMT 2014 English-to-German translation task\cite{mtdataset}.}
	\label{tb:mt}
	\centering
	\begin{tabular}{ll}
		\toprule
		Model     & BLEU \\
		\midrule
		Transformer-base\cite{Vaswani:2017} & 27.3\\
		RoFormer & \textbf{27.5} \\ 
		\bottomrule
	\end{tabular}
\end{table}

\subsection{Pre-training Language Modeling}\label{subsec:pretrain}
The second experiment is to validate the performance of our proposal in terms of learning contextual representations. To achieve this, we replace  the original sinusoidal position encoding of BERT with our RoPE during the pre-training step.

\subsubsection{Experimental Settings} We use the BookCorpus \cite{moviebook} and the Wikipedia Corpus \cite{wikidump} from Huggingface Datasets library (Apache License 2.0) for pre-training. The corpus is further split into train and validation sets at 8:2 ratio. We use the masked language-modeling (MLM) loss values of the training process as an evaluation  metric. 
The well-known BERT \cite{Devlin2019BERTPO} is adopted as our baseline model. Note that we use bert-base-uncased in our experiments.

\subsubsection{Implementation details} 
For RoFormer, we replace the sinusoidal position encoding in the self-attention block of the baseline model with our proposed RoPE and realizes self-attention according to \Cref{fn:rope-qk}. We train both BERT and RoFormer with batch size 64 and maximum sequence length  of 512 for 100k steps. AdamW \cite{adamw} is used as the optimizer with learning rate 1e-5. 

\subsubsection{Results} 
\begin{figure}[hbt]
	\centering
	\includegraphics[width=0.48\textwidth]{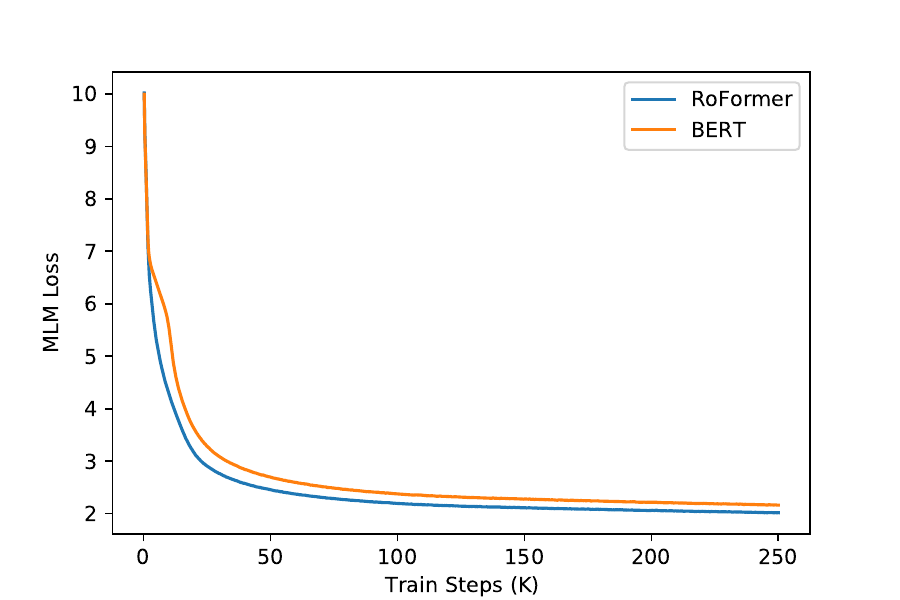}
	\includegraphics[width=0.48\textwidth]{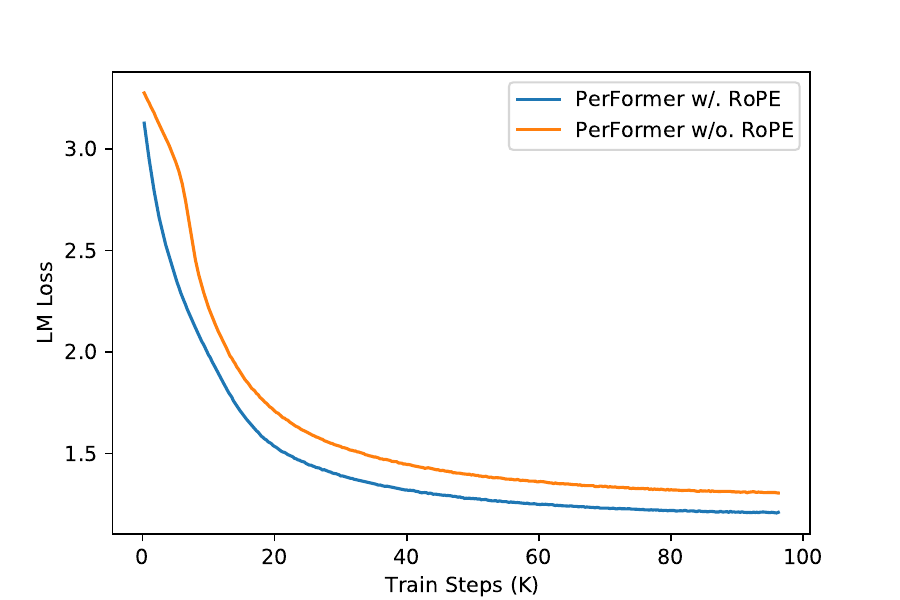}
	\caption{Evaluation of RoPE in language modeling pre-training. \textbf{Left}: training loss for BERT and RoFormer. \textbf{Right}: training loss for PerFormer with and without RoPE.}
	\label{fig:pretrain-loss}
\end{figure}

The MLM loss during pre-training is shown on the left plot of \Cref{fig:pretrain-loss}. Compare to the vanilla BERT, RoFormer experiences faster convergence. 

\subsection{Fine-tuning on GLUE tasks}\label{subsec:glue}
Consistent with the previous experiments, we fine-tune the weights of our pre-trained RoFormer across various GLUE tasks in order to evaluate its generalization ability on the downstream NLP tasks. 

\subsubsection{Experimental Settings} 
We look at several datasets from GLUE, i.e. MRPC \cite{mrpc}, SST-2 \cite{sst2}, QNLI \cite{qnli}, STS-B \cite{stsb}, QQP \cite{qqp} and MNLI \cite{mnli}. We use F1-score for MRPC and QQP dataset, spearman correlation for STS-B, and accuracy for the remaining as the evaluation metrics.

\subsubsection{Implementation details} 
We use Huggingface Transformers library (Apache License 2.0)\cite{wolf-etal-2020-transformers} to fine-tune each of the aforementioned downstream tasks for 3 epochs, with a maximum sequence length of 512, batch size of 32 and learning rates {2,3,4,5}e-5. Following \cite{Devlin2019BERTPO},  we report the best-averaged results on the validation set.
\begin{table}[hbt]
	\caption{Comparing RoFormer and BERT by fine tuning on downstream GLEU tasks.}
	\label{tb:glue-eval}
	\centering
	\begin{tabular}{lcccccc}
		\toprule
		Model & MRPC & SST-2 & QNLI & STS-B & QQP & MNLI(m/mm)\\ 
		\midrule
		BERT\cite{Devlin2019BERTPO} & 88.9 & 93.5 & 90.5 & 85.8 & 71.2 & 84.6/83.4\\
		RoFormer & \textbf{89.5}& 90.7 & 88.0 & \textbf{87.0} & \textbf{86.4} & 80.2/79.8 \\
		\bottomrule
	\end{tabular}
\end{table}
\subsubsection{Results} 
The evaluation results of the fine-tuning tasks are reported in \Cref{tb:glue-eval}. As can be seen, RoFormer can significantly outperform  BERT in three out of six datasets, and the improvements are considerable. 

\subsection{Performer with RoPE}\label{subsec:peformer}
Performer \cite{performer} introduces an alternative attention mechanism, linear attention, which is designed to avoid quadratic computation cost that scales with input sequence length. As discussed in \Cref{sec: prop of RoPE}, the proposed RoPE can be easily implemented in the PerFormer model to realize the relative position encoding while keeping its linearly scaled complexity in self-attention. We  demonstrate its performance with  the pre-training task of language modeling.

\subsubsection{Implementation details}
We carry out tests on the Enwik8 dataset \cite{enwik8}, which is from English Wikipedia that includes markup, special characters and text in other languages in addition to English text. We incorporate RoPE into the 12 layer char-based PerFormer with 768 dimensions and 12 heads\footnote{For this experiment, we adopt code (MIT License) from \url{https://github.com/lucidrains/performer-pytorch}}. 
To better illustrate the efficacy of RoPE,  we report the loss curves of the pre-training process with and without RoPE under the same settings, i.e., learning rate 1e-4, batch size 128 and a fixed maximum sequence length of 1024, etc. 

\subsubsection{Results} 
As shown on the right plot of \Cref{fig:pretrain-loss}, substituting RoPE into Performer leads to rapid convergence and lower loss under the same amount of training steps. These improvements, in addition to the linear complexity,  make Performer more attractive. 

\subsection{Evaluation on Chinese Data}\label{subsec:chinese}
In addition to experiments on English data, we show additional results on Chinese data.  To validate the performance of RoFormer on long texts, we conduct experiments on long documents whose length exceeds 512 characters.

\subsubsection{Implementation} 
In these experiments, we carried out some modifications on WoBERT \cite{zhuiyipretrainedmodels} by replacing the absolute position embedding with our proposed RoPE. As a cross-comparison with other pre-trained Transformer-based models in Chinese, i.e. BERT \cite{Devlin2019BERTPO}, WoBERT \cite{zhuiyipretrainedmodels}, and NEZHA \cite{nezha},  we tabulate their tokenization level and position embedding information in \Cref{tb:roformer-tokpos-imp}.

\begin{table}[h]
	\caption{Cross-comparison between our RoFormer and other pre-trained models on Chinese data. 'abs' and 'rel' annotates absolute position embedding and relative position embedding, respectively. } 
	\label{tb:roformer-tokpos-imp}
	\centering 
	\begin{tabular}{lcccc}
		\toprule
		\makecell[c]{Model} & BERT\cite{Devlin2019BERTPO} & WoBERT\cite{zhuiyipretrainedmodels} & NEZHA\cite{nezha} & RoFormer \\
		\midrule
		Tokenization level & char & word & char & word \\
		Position embedding & abs. & abs. & rel. & RoPE \\ 
		\bottomrule
	\end{tabular}
\end{table}

\subsubsection{Pre-training} 
We pre-train RoFormer on approximately 34GB of data collected from Chinese Wikipedia, news and forums. The pre-training is carried out in multiple stages with changing batch size and maximum input sequence length in order to adapt the model to various scenarios. As shown in \Cref{tb:roformer-pretrain}, the accuracy of RoFormer elevates with an increasing upper bound of sequence length, which demonstrates the ability of RoFormer in dealing with long texts. We claim that this is the attribute to the excellent generalizability of the proposed RoPE.  

\begin{table}[hbt]
	\caption{Pre-training strategy of RoFormer on Chinese dataset. The training procedure is divided into various consecutive stages. In each stage, we train the model with a specific combination of maximum sequence length and batch size. }
	\label{tb:roformer-pretrain}
	\centering
	\begin{tabular}{lccccc}
		\toprule
		Stage & Max seq length & Batch size & Training steps & Loss & Accuracy \\ 
		\midrule
		1 & 512 &  256 & 200k & 1.73 & 65.0\% \\ 
		2 & 1536 & 256 & 12.5k & 1.61 & 66.8\% \\ 
		3 & 256 & 256 & 120k & 1.75 & 64.6\% \\ 
		4 & 128 & 512 & 80k & 1.83 & 63.4\% \\ 
		5 & 1536 & 256 & 10k & 1.58 & 67.4\% \\ 
		6 & 512 & 512 & 30k & 1.66 & 66.2\% \\ 
		\bottomrule
	\end{tabular}
\end{table}

\subsubsection{Downstream Tasks \& Dataset}
We choose Chinese AI and Law 2019 Similar Case Matching (CAIL2019-SCM)\cite{cail2019scm} dataset to illustrate the ability of RoFormer in dealing with long texts, i.e., semantic text matching. CAIL2019-SCM contains 8964 triplets of cases published by the Supreme People's Court of China. The input triplet, denoted as (A, B and C), are fact descriptions of three cases. The task is to predict whether the pair (A, B) is closer than (A, C) under a predefined similarity measure. Note that existing methods mostly cannot perform significantly on CAIL2019-SCM dataset due to the length of documents (i.e., mostly more than 512 characters). We split train, validation and test sets based on the well-known  ratio  6:2:2.

\subsubsection{Results} 
We apply the pre-trained RoFormer model to CAIL2019-SCM with different input lengths. The model is compared with the pre-trained BERT and WoBERT model on the same pre-training data,  as shown in \Cref{tb:roformer-cail2019}. With short text cut-offs, i.e., 512, the result from RoFormer is comparable to WoBERT and is slightly better than the BERT implementation. However, when increasing the maximum input text length to 1024, RoFormer outperforms WoBERT by an absolute improvement of 1.5\%. 

\begin{table}[hbt]
	\caption{Experiment results on CAIL2019-SCM task. Numbers in the first column denote the maximum cut-off sequence length. The results are presented in terms of percent accuracy.}
	\label{tb:roformer-cail2019}
	\centering
	\begin{tabular}{lll}
		\toprule
		\makecell[c]{Model} & Validation & Test \\ 
		\midrule
		BERT-512 & 64.13\% & 67.77\% \\ 
		WoBERT-512 & 64.07\% & 68.10\% \\
		\textbf{RoFormer-512} & 64.13\% & 68.29\% \\
		\textbf{RoFormer-1024} & \textbf{66.07}\% & \textbf{69.79}\% \\ 
		\bottomrule
	\end{tabular}
\end{table}

\subsubsection{Limitations of the work} 
Although we provide theoretical groundings as well as promising experimental justifications, our method is limited by following facts: 
\begin{itemize}
	\item Despite the fact that we mathematically format the relative position relations as rotations under 2D sub-spaces, there lacks of thorough explanations on why it converges faster than baseline models that incorporates other position encoding strategies.
	\item Although we have proved that our model has favourable property of long-term decay for intern-token products, \Cref{sec: prop of RoPE}, which is similar to the existing position encoding mechanisms, our model shows superior performance on long texts than peer models, we have not come up with a faithful explanation.
\end{itemize}
Our proposed RoFormer is built upon the  Transformer-based infrastructure, which requires  hardware resources for pre-training purpose.

\section{Conclusions} \label{sec:conclu}

In this work, we proposed a new position embedding method that incorporates explicit relative position dependency in self-attention to enhance the performance of transformer architectures. Our theoretical analysis indicates that relative position can be naturally formulated using vector production in self-attention, with absolution position information being encoded through a rotation matrix. In addition, we mathematically illustrated the advantageous properties of the proposed method when applied to the Transformer. Finally, experiments on both English and Chinese benchmark datasets demonstrate that our method encourages faster convergence in pre-training. The experimental results also show that our proposed RoFormer can achieve better performance on long texts task.

\bibliographystyle{unsrtnat}
\bibliography{references}  

\end{document}